\documentclass[sigconf]{acmart}
\usepackage{graphicx}
\usepackage{times}
\usepackage{soul}
\usepackage[utf8]{inputenc}
\usepackage{booktabs}
\usepackage{amsmath}
\usepackage{tabu}
\usepackage{makecell}
\usepackage{algorithm}
\usepackage{amsmath}
\usepackage[noend]{algpseudocode}
\usepackage{algpseudocode}
\usepackage{stfloats}
\usepackage{booktabs}
\usepackage{bbm}
\usepackage{bbold}
\usepackage[utf8]{inputenc}
\usepackage{multirow}
\usepackage{subfigure}
\usepackage{adjustbox}
\algnewcommand\algorithmicforeach{\textbf{for each:}}
\algnewcommand\ForEach{\item[ \algorithmicforeach]}
\usepackage{graphicx} 
\usepackage{xcolor}
\usepackage{algorithm}
\usepackage{tabularx} 
\usepackage{pdfpages}
\AtBeginDocument{%
  \providecommand\BibTeX{{%
    \normalfont B\kern-0.5em{\scshape i\kern-0.25em b}\kern-0.8em\TeX}}}

\setcopyright{acmcopyright}
\copyrightyear{2020}
\acmYear{2020}
\acmDOI{10.1145/1122445.1122456}

\acmConference[Wisdom@KDD '20]{Wisdom '20: Workshop on Issues of Sentiment Discovery and Opinion Mining}{August , 2020}{SAN DIEGO, CA}

\begin{document}

\title{SentiQ: A Probabilistic Logic Approach to Enhance Sentiment Analysis Tool Quality}

\author{Wissam Maamar Kouadri}
\email{wissam.maamar\_kouadri@u-paris.fr}
\affiliation{%
  \institution{Université de Paris}
}

\author{Salima Benbernou}
\email{salima.benbernou@u-paris.fr}
\affiliation{%
  \institution{Université de Paris}
}

\author{Mourad Ouziri}
\email{mourad.ouziri@u-paris.fr}
\affiliation{%
  \institution{Université de Paris}
}

\author{Themis Palpanas}
\email{themis.palpanas@u-paris.fr}
\affiliation{%
  \institution{ Université de Paris \& Institut Universitaire de France (IUF)}
  \
}
\author{Iheb Ben Amor}
\email{Iheb.Benamor@imbaconsulting.fr}
\affiliation{%
  \institution{IMBA Consulting}
  \
}

\renewcommand{\shortauthors}{Maamar kouadri et al.}

\begin{abstract}
  The opinion expressed in various Web sites and social-media is an essential contributor to the decision making process of several organizations. 
Existing sentiment analysis tools aim to extract the polarity (i.e., positive, negative, neutral) from these opinionated contents. 
Despite the advance of the research in the field, sentiment analysis tools give \textit{inconsistent} polarities, which is harmful to business decisions. 
In this paper, we propose SentiQ, an unsupervised Markov logic Network-based approach that injects the semantic dimension in the tools through rules. 
It allows to detect and solve inconsistencies and then improves the overall accuracy of the tools. 
Preliminary experimental results demonstrate the usefulness of SentiQ.

\end{abstract}






\maketitle

\section{Introduction}

With the proliferation of social media,  people are increasingly sharing their sentiments and opinions online about products, services, individuals, and entities, which has spurred a growing interest in sentiment analysis tools in various domains \cite{greene2009more,dragoni2018fuzzy,feldman2013techniques,gilbert2014vader,kouloumpis2011twitter,DBLP:conf/www/WangSH0Z18,DBLP:journals/tkde/TsytsarauP16}. The customer opinion, if yielded correctly, is crucial for the decision-making of any organization. 
Thus, numerous studies~\cite{socher2013recursive,cambria2018senticnet,kim2014convolutional,gilbert2014vader,severyn2015twitter}
try to automate sentiment extraction from a massive volume of data by identifying the polarity of documents, i.e., positive,  negative, or neutral.

Nevertheless, sentiment analysis of social media data is still a challenging task\cite{farias2017irony} due to the complexity and variety of natural language through which the same idea can be expressed and interpreted using different text. 
Many research work have adopted the consensus that semantically equivalent documents should have the same polarity 	\cite{cambria2018senticnet,ding2018weakly,fu2014improving,vosoughi2016tweet2vec,wei2019eda,risch2018aggression}. For instance \cite{ding2018weakly} have attributed the same polarity labels to the semantically equivalent couples (event/effect) while  \cite{fu2014improving} have augmented  their sentiment dataset using paraphrases and assign the original document's polarity to the generated paraphrases.

However, we found that these tools do not detect this similarity and assign different polarity labels to semantically equivalent documents; hence, considering \textit{in-tool } inconsistency where the sentiment analysis tool attribute different polarities to the semantically equivalent documents and \textit{inter-tool inconsistency} where different sentiment analysis tools attribute different polarities to the same documents that have a single polarity. This inconsistency can be translated by the fact that at least one tool has given an incorrect polarity. Consequently,  returning an incorrect polarity in the query  can be misleading, and leads to poor business decision.

Few works have used inconsistencies to improve systems' accuracy, such as \cite{DBLP:journals/vldb/RatnerBEFWR20}, that considers various labeling functions and  minimizes the inter-tool inconsistency between them based on different factors: correlation, primary accuracy, and labelling abstinence. However, in \cite{DBLP:journals/vldb/RatnerBEFWR20}, we resolve the inconsistency  statistically, and ignore the semantic dimension that could enhance the results' quality. The work in \cite{ding2018weakly} has proposed to create a corpus of (event/effect) pairs for sentiment analysis by minimizing the sentiment distance between semantically equivalent (event/effect) pairs. In our work, we study the effect of solving the two types of inconsistency on  accuracy. We focus more on the improvement that we can  obtain by resolving in-tool inconsistency between the documents i.e., resolving inconsistency such that all semantically equivalent documents get the same polarity label and resolving both inconsistencies. 
To the best of our knowledge, the only work studying polarity inconsistency does this at word-level~\cite{DBLP:journals/tkde/DragutWSYM15}, by checking the polarity consistency for sentiment words inside and across dictionaries.

Our work is the first to study the effect of resolving the polarity inconsistency on accuracy for in-tool inconsistency,  \emph{and} inter-tool inconsistency on document data. We seek to converge to the golden truth by resolving in-tool and inter-tool inconsistencies. Each document has a unique polarity, by resolving in-tool and inter-tool inconsistency, we minimize the gap of incorrect labels and converge to the gold truth. Such a method can be applied on any classification task in natural language processing. 

\noindent{\bf Contributions.}
In summary, we make the following contributions:
\begin{itemize}
\item 
We study the impact of inconsistency on the accuracy of the sentiment analysis tools.
\item
We propose SentiQ, an approach that resolves both polarity inconsistencies: in-tool and inter-tool. The approach we are proposing is based on our earlier work to handle the inconsistency in big data~\cite{benbernou2017enhancing} on one side and on the probabilistic logic framework, Markov Logic Network, on the other side.
\item
We present preliminary experimental results using news headlines datasets~\cite{cortis2017semeval} and the sentiment treebank dataset~\cite{socher2013recursive}. 
When compared to  the majority voting to resolve  inter-tool inconsistencies, our framework leads to the efficiency of using the semantic dimension in optimizing the accuracy by resolving both in-tool and inter-tool inconsistencies. 
\item
Following the lessons learned from our experimental evaluation, we discuss promising future research directions, including the semantic dimension's use in different integration problems, such as truth inference in crowd-sourcing and accuracy optimization of different classification problems. \end{itemize}

\noindent{\bf Paper Outline.}
In the remainder of the paper, we present in section 2 a motivation through a real example. In section 3, we provide some preliminaries used in our work. In sections 4 and 5, we discuss the SENTIQ model based on Markov Network logic (MLN) while in section 6, we present our experiments and discussions.





\section{Motivating Example}
We consider the following real life example collected from twitter and that represents statements
about Trump's restrictions on Chinese technology such that $ D= \{d_1,\dots , d_9 \}$ and:
\begin{itemize}
\item \textbf{$d_1$} : Chinese technological investment is the next target in Trump's crackdown. 
\item\textbf{$d_2$} : Chinese technological investment in the US is the next target in Trump's crackdown.
\item\textbf{$d_3$} : China urges end to United States crackdown on Huawei.
\item\textbf{$d_4$} : China slams United States over unreasonable crackdown on Huawei. 
\item\textbf{$d_5$} : China urges the US to stop its unjustifiable crackdown on Huawei.
\item\textbf{$d_6$} : Trump softens stance on China technology crackdown. 
\item\textbf{$d_7$} : Donald trump softens threat of new curbs on Chinese investment in American firms. 
\item\textbf{$d_{8}$} : Trump drops new restrictions on China investment.
\item \textbf{$d_{9}$} : Donald Trump softens tone on Chinese investments.
\end{itemize}
We call each element of this dataset $D$ a $Document$. 
We notice that $D$ can be clustered on subsets of semantically equivalent documents. For instance, $d_1$ and $d_2$ are semantically equivalent as they both express the idea that the US is restricting Chinese technological investments. 
We denote this set by $A_1$ and we write: $A_1=\{d_1,d_2\}$ and $A_2=\{ d_3, d_4, d_5\}$, 
which express that the Chinese government demands the US to stop the crackdown on Huawei, and $A_3=\{d_6, \dots , d_{9} \}$
which conveys the idea that Trump reduces restrictions on Chinese investments. 
We have: $D= A_1 \cup A_2 \cup A_3$. 
We analyse $D$ using three sentiment analysis tools: Stanford Sentiment Treebank~\cite{socher2013recursive}, Sentiwordnet~\cite{baccianella2010sentiwordnet} and Vader~\cite{gilbert2014vader}.
In the rest of this paper, we refer to the results of these tools using the polarity functions: $P_{tb}$,  $P_{sw}$, $P_{v}$; we use $P_h$ to refer to the ground truth. 
Table~\ref{tab:t3} summarizes the results of the analysis.
	\begin{table}[tb]
		\centering
		{%
			\begin{tabu}{ c c c c c c  } 
				\hline
				$A_i$ &Id & $P_{tb}$ &  $P_{sw}$ & $P_{v}$ & $P_h$ \\ 
				\hline
				$A_1$ &$d_1$&	Neutral &	Negative&	Neutral	& Negative \\
				&$d_2$&	Negative&Negative&	Neutral &	Negative\\
				\hline
				$A_2$ &$d_3$&	Negative&	Positive&	Neutral&	Negative\\
				&$d_4$&	Negative	&	Negative&	Neutral&	Negative\\
				&$d_5$&	Negative&	Negative&	Neutral& Negative\\
				\hline
				$A_3$ &$d_6$ &	Neutral &		Positive&	Neutral	 &Positive\\
				&$d_7$ &	Negative &		Negative&	Negative	&Positive\\
				&$d_{8}$ &	Negative&	 Positive&	Neutral &Positive\\
				&$d_{9}$ &	Neutral &	Negative&	Neutral &Positive\\
				\hline
		\end{tabu}}
		\caption{{Predicted polarity on dataset D by different tools}}
		
		\label{tab:t3}
		
	\end{table}
	
We know that each document has a single polarity, so each precise tool should find this polarity, and a difference in prediction results is a sign that at least one tool is erroneous on this document.  We also know that semantically equivalent documents should have the same polarity. However, in this real-life example, we observe different tools attributing different polarities for the same document (e.g., only $P_{tb}$ attributes the correct polarity to $d_3$ in $A_2$), which represent an inter-tool inconsistency. 
 Also, the same tool attributes different polarities for semantically equivalent documents (for e.g., $P_{tb}$ considers $d_6$ as Neutral and $d_7$ as Negative) which represent an in-tool inconsistency.  A trivial method to resolve those inconsistencies is to use majority voting, inside the cluster of documents, or between functions. However, when applying the majority voting baseline on this example, we found that the polarity is $Negative$ in $A2$ which represents the correct polarity of the cluster while we found that the polarity  is   $Negative$ in $A_3$, which is not a correct polarity in this case.  Because with simple majority voting, we got only a local vision of the polarity function,  and we ignore its behavior on the rest of the data. 
	
	\section{Preliminaries}
	\begin{definition}{{(Sentiment Analysis)}}
		\end{definition}
	\noindent
	  \textbf{Sentiment Analysis} is the process of extracting a polarity $\pi \in \{+,-,0\}$ from a document $d_i$. With  {$+$} for Positive polarity, {$-$} for Negative polarity and {$0$} for Neutral polarity. In this paper, we refer to polarity functions as $P_{t_k}$ s.t: $P_{t_k}:D \to \pi$. We refer to the set of all functions as $\Pi$ s.t $\Pi = \{P_{t_1}, \dots , P_{t_n}\}$
	 
	 \noindent
	 
	 \noindent
	 
	 \noindent
	 	\begin{definition}{{(Polarity Consistency)}}
		\end{definition}
	\noindent
	 \textbf{Cluster:} cluster is a set of semantically equivalent documents: 
	 
	 \noindent
	 	for a cluster $A_l=\{d_1,\dots,d_n\} \, \text{ we have } \forall d_i\,, d_j \in A_l \,\,$  $, d_i \stackrel{s}{\iff}\,\,d_j $. \\

	 \noindent
	 \textbf{Sentiment Quality:}
	 we define the polarity consistency of a given cluster $A_i$ as the two following rules:
	 
	 \noindent 
	 \textbf{In-tool Consistency} means that semantically equivalent documents should get the same polarity, s.t.:
	 \begin{equation}
	     \forall d_i,d_j \in A, P_{*} \in \Pi \, P_{*}(d_i) =P_{*}(d_j) 
	     \label{eq:in-tool-cons}
	 \end{equation}

	 \noindent
    	 \textbf{Inter-tool Consistency} means that all polarity functions should give the same polarity to the same document:
	 \begin{equation}
	 \forall d_i \in A, P_{t_k},P_{t_k'}  \in \Pi \, P_{t_k}(d_i) =P_{t_k'}(d_i) \label{eq:intertool-cons}
	 \end{equation}
	 
	\begin{definition}{{(Markov Logic Network (MLN))}}
		\end{definition}
	\noindent

In this section, We recall  Markov logic network	(MLN) model \cite{DBLP:journals/ml/RichardsonD06,DBLP:journals/cacm/DomingosL19}  which is a general framework for joining logical and Probability.

\noindent
\textbf{MLN} is defined as a set of weighted first-order logic (FOL) formula with free variables $L= \{(l_1, w_1),\dots, (l_n,w_n)\}$, with $w_i \in I\!R \cup \infty$ and $l_i$ an FOL constraint. With a set of constants $C=\{c_1, \dots, c_m\}$, it constitutes the Markov network $M_{L,C}$. The $M_{L,C}$ contains one node for each predicate grounding that its value is 1 if the grounding is true and 0 otherwise. Each formula of $L$ is represented by a feature node that its value is 1 if the formula $l_i$ grounding is true and 0 otherwise.  The syntax of the formulas that we adopted in this paper is the FOL syntax. 

\noindent 
\textbf{World} $x$ over a domain $C$ is a set of possible grounding of $MLN$ constraints over $C$.

\noindent
\textbf{Hard Constraints} are constraints with infinite weight $w_i=\infty$. A world $x$ that violates these constraints is impossible. 

\noindent
\textbf{Soft Constraints} are  constraints with a finite weight ($w_i \in I\!R) $ that can be violated. 


\noindent
\textbf{World's Probability} is the probability distribution of possible worlds $x$ in $M_{L,C}$ given by 
$$Pr(X=x) = \frac{1}{Z} exp(\sum_i{w_i,n_i(x))}$$, where $n_i(x)$ is the number of the true grounding of $F_i$ in  $x$ and $Z$ is a normalization factor.

\noindent
\textbf{Grounding.}
We define grounding as the operation of replacing  predicate variables  by constants from $C$.

\section{SentiQ: An MLN based model for inconsistency}
The polarity inconsistency is a complex problem due to the tool and document natures and the relations between them. 
This problem can be solved using semantics to model the relations between tools, documents, and statistic dimension to optimize both the inconsistency and the accuracy of the system 
—this why we chose $MLN$ to model the resulted inconsistent system.
 We present the details of our semantic model in this section.


\subsection{Semantic Model's Components}

Our semantic model is a knowledge-base $KB=<R,F>$, where
\begin{enumerate}
\item $R$ is a $\textit{ set of rules}$  (FOL formulas) defining the vocabulary of our application which consists of concepts (sets of individuals) and relations between them. 

\item $F$ is a $\textit{ set of facts}$   representing the instances of the concepts or individuals defined in $R$ .
\end{enumerate}

We represent each document by the concept $Document$, each polarity function in the system by its symbol and the polarity that it attributes to the $Document$. For instance, $P_{tb+(d_1)}$, $P_{tb0}$, and $P_{tb-}$ represent respectively the polarities (+, 0, -) attributed to the $Document(d_1)$ by the polarity function $P_{tb}$. Each $Document$ is  $Positive$, $Negative$, or $Neutral$. This is represented respectively by the concepts $IsPositive$, $IsNegative$, and $IsNeutral$.
 We also have  the relation $sameAs$ as a semantic similarity  between documents in the input dataset clusters. For instance, $sameAs(d_1,d_2)$ indicates that the documents $Document(d_1)$ and $Document(d_2)$ are semantically equivalent. 


\subsection{Rule modeling for inconsistency}

We define two types of rules from $R$ in our framework, \textit{Inference rules} and \textit{Inconsistency rules}:

\textit{Inference rules IR}

The inference rules allow deriving the implicit instances. They model the quality of the polarity at in-tool and inter-tool levels. They are soft rules that  add an uncertainty layer to different polarity functions based on the inconsistency of tools. 

\noindent
$\bullet$ \textbf{In-tool consistency rules.}
This set of rules models the fact that all the documents of the cluster should have the same polarity. They are defined as follows (for the sake of clarity we omitted the predicate $Document(d_i)$ in all logical rules) :

\begin{flalign*}
    IR1:   sameAs(?d_i,?d_j)\land IsPositive(?d_j) \to IsPositive(?d_i) \\
IR2:  sameAs(?d_i,?d_j)\land IsPositive(?d_i) \to IsPositive(?d_j)
\end{flalign*}

\noindent

The rule $IR1$ denotes that if two documents $d_i$ and $d_j$ are semantically equivalent (expressed with $sameAs$ relation), they got the same polarity, which translates the in-tool consistency defined in equation~\ref{eq:in-tool-cons}.
The $sameAs$ relation is transitive, symmetric, and reflexive. We express the symmetry by duplicating the rule for both documents of the relation (rules $IR1$ and $IR2$ instead of only one rule). For instance, when applying the rule $IR1$ on the relation $sameAs(d1,d2)$ and the instances $IsNeutral(d1)$ and $IsNegative(d2)$, we infer the new instance $IsNeutral(d2)$. The instance $IsNegative(d1)$ is inferred when applying the rule $IR2$. 
The transitivity is handled in the instantiating step (algorithm~\ref{alg:instatiation}) and we ignore the  reflexivity  of the relation because it does not infer additional knowledge. Note that $IR1$ and $IR2$ are examples of rules. The set of rules is presented in Algorithm~\ref{alg:iplicit-inf}.
 
\noindent
$\bullet$ \textbf{Inter-tool consistency rules.}
These rules model the inter-tool consistency described in equation~\ref{eq:intertool-cons} by assuming that each function gives the correct polarity to the document.  For example, given the instances $P_{tb-}(d2)$ the rule $IR$ infers $IsNegative(d2)$. For each tool in the system, we create the following rules by replacing $P_{t_k*}$ with the polarity function of the tool.

\begin{flalign*}
IR3: &P_{t_k+} (?d_i) \to IsPositive(?d_i) \,\,\, IR4:P_{t_k-} (?d_i) \to IsNegative(?d_i) \\
IR5: &P_{t_k0} (?d_i) \to IsNeutral(?d_i)
\end{flalign*}
Those rules are soft rules that allow us to represent inconsistencies in the system and attribute a ranking to the rules that we use in the in-tools uncertainty calculation. 
The idea behind this modeling is that if the inter-tool consistency is respected, all tools will attribute the same polarity to this document; otherwise, the document will have different polarities  (contradicted polarities). To represent this contradiction, we define, next, inconsistency rules.





 \textit{Inconsistency rules ICR}
 
 They are considered as hard rules that represent the disjunction between polarities since each document has a unique polarity. 
 $$ ICR1: IsPositive(?d_i) \to  \neg IsNegative(?d_i) \land  \neg IsNeutral(?d_i) $$
 $$ ICR2: IsNegative(?d_i) \to  \neg IsPositive(?d_i) \land  \neg IsNeutral(?d_i) $$
 $$ ICR2: IsNeutral(?d_i) \to  \neg IsPositive(?d_i) \land  \neg IsNegative(?d_i) $$

 These rules generate negative instances that create inconsistencies used in learning inference rules weights. 
 
 \noindent
For instance, consider the following instances $P_{tb^{-}}(d_3)$ and $P_{sw^{+}}(d_3)$ from the motivating example. 
 By applying the inter-tool consistency inference rules, we infer: $IsNegative(d_3)$ and $IsPositive(d_3)$. However, F appears consistent even it contains polarity inconsistencies. We get the inconsistency once applying the inconsistency rules. We get:  $\neg IsPositive(d_3)$, $\neg IsNeutral(d_3) $, $\neg IsNegative(d_3)$  that represent an apparent inconsistency in F.
 
 \subsection{MLN based model for inconsistency resolution}
 As depicted in Figure \ref{fig:inconistency}, the proposed inconsistency resolution process  follows  four main phases:
 \begin{itemize}
     \item Inference of implicit knowledge: The system infers all implicit knowledge needed for the inconsistencies  before applying the learning procedure of $MLN $.
     \item Detection of inconsistencies: Having the explicit and implicit knowledge, we apply the inconsistency rules to discover the inconsistency. 
     \item Inference of correct polarity: Using the saturated fact set  $F$ and $R$, the system learns first the weights of $M_{R,F}$, and use them to infer the correct polarities. 

     \item Resolve the in-tool inconsistencies : Since we are in an uncertain environment, we can still have some in-tool inconsistencies after the previous phase, that we resolve by applying a weighted majority voting. 
 \end{itemize}
 The phases will be detailed in the next section. 
\section{SENTIQ:The inconsistency resolution}
In this section we discuss the reasoning process to solve the inconsistencies and improve the accuracy.
 \subsection{Facts generation }
 
Our data are first saved in a relational database, where each table represents a concept, and the table content represents the concept's domain.  For that, instantiating our data follows the steps of Algorithm \ref{alg:instatiation}.
  
   Each function and its polarity is represented by a table. The content of the table is the document ID  that got this polarity by the function. The instantiating process converts the content of the database to logic predicates that we use in our reasoning. 
  The purpose of this algorithm is to fill in the set $F$ with the prior knowledge needed in the reasoning.  Our prior knowledge is the documents, polarities attributed by the functions to documents, and the semantic similarity between documents represented by the $SameAs$ predicate. 
  We note that we do not consider the ground truth. We adopt an unsupervised approach because inconsistency resolution is useful when we do not know the valid prediction from the invalid ones.
 
 \begin{algorithm}
		\caption{Instantiating}
		\begin{flushleft}
			\hspace*{\algorithmicindent} \textbf{Input} : $\text{Database with prior knowledge}$ \\
			\hspace*{\algorithmicindent} \textbf{Output} : $\text{F:Set of generated Facts (polarities and same as)}$
		\end{flushleft}
		\begin{algorithmic}[1]
			\Procedure{Instantiating}{}

			
			\State \hspace{0.5cm}$\textbf{//Step1: }\text{Add all Polarities attributed to documents}$
			\State \hspace{0.5cm} $\textbf{for each } P_{t_k} \in \textit{Functions}:$
			\State \hspace{1cm} $\textbf{for each } d_i \in P_{t_k^+}:\textit{F.add}(P_{t_k^+}(d_i))$
			\State \hspace{1cm} $\textbf{for each } d_i \in {P_{t_k^-}}:\textit{F.add}(P_{t_k^-}(d_i))$
			\State \hspace{1cm} $\textbf{for each } d_i \in {P_{t_k^0}}:\textit{F.add}(P_{t_k^0}(d_i))$
			
			\State \hspace{0.5cm}$\textbf{//Step2: }\text{Add sameAs relations}$

			\State \hspace{0.5cm} $\textit{clusters =  groupeByClusterId(D)}$
			\State \hspace{0.5cm} $\textbf{for each } cluster \in \textit{clusters}:$
			\State \hspace{1cm} $\textbf{for} i \in \textit{\{0,\dots, len(cluster)\}}:$
			\State \hspace{1.5cm} $\textbf{for} j \in \textit{\{i+1 ,\dots, len(cluster)\}}:$
			\State \hspace{2cm} $\textbf{if }  SameAs(d_i,d_j) \notin \textit{F}:$
			\State \hspace{2cm}$\textit{F}.add(SameAs(d_i,d_j))$
			\State $\textbf{return } F$
			\EndProcedure
		\end{algorithmic} 
		\label{alg:instatiation}
	\end{algorithm}

  \subsection{Implicit knowledge inference Algorithm}
In $MLN$, the  learning is done only on the available knowledge in $F$. For this, we infer all implicit knowledge in the system before applying the learning procedure. The inference procedure is presented in Algorithm~\ref{alg:iplicit-inf}. This inference phase is crucial  for an integrated learning since most polarity knowledge are implicit. For instance, consider the two documents $d_3$ and $d_4$ from the motivating example. We have $P_{sw+}(d_3)$ and $P_{sw-}(d_4)$, by inferring documents polarities using inter-tool consistency rules $IR3$ and $IR4$, we get $IsPositive(d_3)$ and $IsNegative(d_4)$. When applying the in-tool consistency rules on the previous concepts and the relation $sameAs(d_4,d_3)$ ($IR1$ and $IR2$), we infer the new polarities $IsNegative(d_3)$ and  $IsPositive(d_4)$.
 
   We ensure that we inferred all implicit knowledge by redoing the inference until no new knowledge are inferred. Such process is called inference by saturation.  
     
  \begin{algorithm}
		\caption{Implicit Knowledge Inference Algorithm}
		\begin{flushleft}
			\hspace*{\algorithmicindent} \textbf{Input} : $F~ with ~ prior ~ knowledge$ \\
			\hspace*{\algorithmicindent} \textbf{Output} : $Saturated ~ F$
		\end{flushleft}
		\begin{algorithmic}[1]
			\Procedure{Inference}{}
			\State \hspace{0.25cm}$\textbf{//step1: } \text{Infer Polarities by applying}$ \State \hspace{0.25cm}$\textbf{//}\text{inter-tool consistency rules}$
			\State \hspace{0.25cm} $\textit{ functions=D.getFunctions( )} d_i$
			\State \hspace{0.25cm} $\textbf{for each} P_{t_k^+}(d_i)  \in \textit{functions}:$
			\State \hspace{0.5cm} $\textbf{if } {IsPositive(d_i)} \notin {F:    F.add(IsPositive(d_i))}$
			\State \hspace{0.25cm} $\textbf{for each} P_{t_k^-}(d_i)  \in \textit{functions}:$
			\State \hspace{0.39cm} $\textbf{if } {IsNegative(d_i)} \notin {F:}$
			 $F.add(IsNegative(d_i))$
			 			\State \hspace{0.25cm} $\textbf{for each} P_{t_k^0}(d_i)  \in \textit{functions}:$
			\State \hspace{0.65cm} $\textbf{if } {IsNeutral(d_i)} \notin {F:    F.add(IsNeutral(d_i))}$

						\State \hspace{0.25cm}$\textbf{//step2: } \text{Infer Polarities by applying}$
\State \hspace{0.25cm}$\textbf{//}\text{in-tool consistency rules}$
			\State \hspace{0.25cm} $\textbf{sameAsRelations} = getSameAs(F)$
			\State \hspace{0.25cm} $\textbf{repeat:}$
			\State \hspace{0.65cm} $\textbf{for each } \textit{SameAs} \in sameAsRelations:$
			\State \hspace{0.9cm} $\textbf{if }IsPositive(d_i)\in F\land IsPositive(d_j)\notin F:$ 
			\State \hspace{1.2cm}${F.add(IsPositive(d_j))}$
            \State \hspace{0.9cm} $\textbf{if }IsPositive(d_j) \in F \land IsPositive(d_i) \notin F:$ 
            \State \hspace{1.2cm}${F.add(IsPositive(d_i))}$
            \State \hspace{0.75cm} $\textbf{if }IsNegative(d_i) \in F \land IsNegative(d_j) \notin F:$
            \State \hspace{1.2cm}${F.add(IsNegative(d_j))}$
            \State \hspace{0.75cm} $\textbf{if }IsNegative(d_j) \in F \land IsNegative(d_i) \notin F:$
            \State \hspace{1.2cm}${F.add(IsNegative(d_i))}$
            \State \hspace{0.9cm} $\textbf{if }IsNeutral(d_i) \in F\land IsNeutral(d_j) \notin F:$
            \State \hspace{1.2cm}$ {F.add(IsNeutral(d_j))}$
            \State \hspace{0.9cm} $\textbf{if }IsNeutral(d_j) \in F \land IsNeutral(d_i) \notin F:$ 
            \State \hspace{1.2cm}${F.add(IsNeutral(d_i))}$
            \State \hspace{0.25cm}$\textbf{until: }\textit{no new inferred instance}$
            \State \hspace{0.25cm}$\textbf{return: }\textit{F}$
			\EndProcedure
		\end{algorithmic} 
		\label{alg:iplicit-inf}
	\end{algorithm}

 \subsection{Inconsistency inference Algorithm}
After inferring all implicit knowledge in the set $F$, we apply the inconsistency rules $ICR$ that allow to explicitly define the inconsistencies as it is presented in Algorithm~\ref{alg:inc-inf}. We apply this rules on a saturated knowledge base because most inconsistencies are implicit. For instance, if we apply the inconsistency rules directly after inferring the polarities $IsPositive(d_3)$ and $IsNegative(d_4)$,  we get $\neg IsNegative(d_3)$, and $\neg IsPositive(d_3)$. However, when applying the in-tool consistency rules on the previous concepts and  relation $sameAs(d_4,d_3)$ (saturation process), we obtain $IsNegative(d_3)$ and  $IsPositive(d_4)$, and when applying the inconsistency rules on this instances, we get $\neg IsPositive(d_3)$ and $\neg IsNegative(d_4)$ which represents an implicit inconsistency in the fact set $F$. Therefore, applying the inconsistency rules on $F$ after the saturation process is an important step in our reasoning procedure, because it shows all inconsistencies even the implicit ones.

\begin{figure}[ht]
		\centering
		\includegraphics[ width=0.5\textwidth, ]{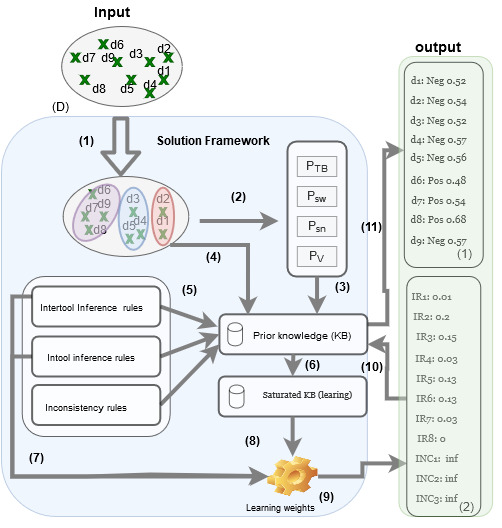}
		\caption{SentiQ overview 
		}
		\label{fig:flow}
		\label{fig:inconistency}
	\end{figure}

\subsection{MLN reasoning to reduce inconsistency}
Here we discuss how to reduce  the inconsistencies discovered in the previous phase, by applying the $MLN$ approach. The reasoning process will first learn the rules' weights of $R$ and after will use them to infer the correct polarities.\\

\noindent 
\textbf{Grounding.} The grounding algorithm enumerates all possible assignments of formulas to its free variables. (the set of possible worlds). We used the grounding algorithm described in~\cite{niu2011tuffy} because it speeds up the inference process. We adopted the closed world assumption; hence we consider all groundings that are not present in the Fact set as false.

\noindent 
\textbf{Learning.} To learn the rules' weights, we use the discriminative training described in \cite{lowd2007efficient}. The training consists of optimizing  the  conditional log-likelihood given by:    

$$
\log{-P(Y=y| X=x)} = \log {Z_x} - \sum_i w_i n_i (x,y) 
$$
where  $X$ represents priors (saturated inconsistent fact set), Y the set of queries (in our case: $IsPositive(d)$, $IsNegative(d)$, $IsNeutral(d)$),  $Z_x$ the normalization factor over the set of worlds, and $n_i(x,y)$ the number of the correct groundings of the formula $l_i$ (the inference rules) in the worlds where $Y$ holds. 

We used in the optimization the Diagonal Newton discriminative method described in~\cite{lowd2007efficient} that calculates the $Hessain$ of the negative conditional log-likelihood given by: 
$$
\frac{\partial}{\partial w_i \partial w_j} - log P(Y=y| X=x)  = E_w[n_in_j]-E_w[n_i]E_w[n_j]
$$

With $E_w$ the expectation.  We call the inference procedure MC-SAT to estimate the number of satisfied (correct) formulas ($n_i$, $n_j$). 

 We can see that we consider the rules independently in the learning process. We calculate the number of each formula's correct grounding separately in the world; hence we do not take into consideration the implicit knowledge, which justifies the inference of all implicit knowledge and inconsistencies before learning. 

\noindent 
\textbf{Inference.}
The inference in $MLN$ \cite{DBLP:journals/ml/RichardsonD06} contains two steps, grounding step, where we sample all possible worlds based on the priors and construct a  large weighted Sat formula used in satisfiability calculation,   and search step to find the best weight assignment to this Sat formula.  In our work, we used the  marginal inference algorithm  that estimates the atoms' probability and returns the query answer with a probability score representing the confidence. It uses the MC-Sat algorithm, which combines satisfiability verification with MCMC by calling in  each step the SampleSat algorithm that is a combination of Simulated Annealing and WalkSat. Note that the walkSat algorithm selects in each iteration an unsatisfiable clause, selects an atom from the clause, and flip its truth value to satisfy the clause.

\subsection{MLN-based Reasoning to Enhance Accuracy}

Majority voting could be a solution to the inconsistency problem. 
However, this trivial method takes into consideration only the voting subset (cluster) and ignores  information about the voters (polarity functions) from the other voting subsets (other clusters), which may hurt the accuracy.

To enhance the quality in terms of accuracy of the inconsistency issue resolution, the process in SentiQ follows two steps: 

\textbf{Step1.} We use $MLN$ to model the different inconsistencies and select the most appropriate polarity of the set (phase 1 to phase 3 of the process). 
We illustrate in Figure~\ref{fig:flow}  the global workflow of our system.
As input, we have an unlabeled dataset $D$ (1) that we cluster to group the semantically equivalent documents in clusters. 
Then, (2) we extract the polarities from the documents using different polarity functions ($P_{tb}$, $P_{sw}$,  $P_{v}$). 
After that, (4) we construct our knowledge base $KB$ by creating first the fact set $F$ (Algorithm~\ref{alg:instatiation}).
 (5) We infer all implicit knowledge by applying inference rules ($IR$)  on the \textit{Fact set} until saturation( Algorithm~\ref{alg:iplicit-inf}). Then we apply inconsistency rules (ICR) to generate different inconsistencies between polarities (Algorithm~\ref{alg:inc-inf}).
 (7) We learn the weights of  inference rules. 
(8) The output of the learning procedure is a set of weighted inference rules that we apply on the\textit{prior knowledge} to infer the most appropriate polarities for documents.

Running the motivating example in this system shows an improvement in both the consistency and accuracy (accuracy of 88\% and a low inconsistency).

\textbf{Step2.} (phase 4 of the process) As we still have inconsistencies from the previous step, we propose to resolve those remaining inconsistencies by using weighted majority voting with as weights the polarities probability, which leads to an accuracy of 100\% on the motivating example. 

  \begin{algorithm}
		\caption{Discover inconsistencies}
		\begin{flushleft}
			\hspace*{\algorithmicindent} \textbf{Input} : $Saturated ~ F$ \\
			\hspace*{\algorithmicindent} \textbf{Output} : $Inconsistent ~ F$
		\end{flushleft}
		\begin{algorithmic}[1]
			\Procedure{Inconsistency Inference}{}
			\State \hspace{0.25cm}$\textbf{//step1: }\text{get all polarities from the F}$
			\State \hspace{0.25cm}$\textbf{//}\text{and apply inconsistency rules}$
			
			\State \hspace{0.25cm} $\textit{polarities=F.getPolarities()}$
            \State \hspace{0.25cm} $\textbf{for each  } Polarity \in polarities:$
            \State \hspace{0.4cm} $\textbf{if  } Polarity == IsPositive(d_i):$
            \State \hspace{1cm}$F.add( \neg IsNegative(d_i) )$
            \State \hspace{1cm} $ F.add( \neg IsNeutral(d_i) )$
            \State \hspace{0.4cm} $\textbf{if  } Polarity == IsNegative(d_i): $
            \State \hspace{1cm}$F.add( \neg IsPositive(d_i) )$
            \State \hspace{1cm}$F.add( \neg IsNeutral(d_i) )$

            \State \hspace{0.5cm} $\textbf{if  } Polarity == INeutral(d_i) :$
            \State \hspace{1cm} $F.add( \neg IsPositive(d_i) )$
            \State \hspace{1cm}$F.add( \neg IsNegative(d_i) )$
		
			\EndProcedure
		\end{algorithmic} 
		\label{alg:inc-inf}
	\end{algorithm}	
	\section{Experimental Evaluation}
	
	\textbf{Tools.}
	In our experiments, we use five representative sentiment analysis tools, a convolutional  neural network with word embedding as $P_{cnn\_txt}$\cite{kim2014convolutional},  a convolutional  neural network with character embedding as $P_{char\_cnn}$\cite{dos2014deep} , \cite{gilbert2014vader} as $P_v$, \cite{baccianella2010sentiwordnet} as $P_{sw}$ and \cite{socher2013recursive} as $P_{tb}$.  We chose these tools because of their performance and their association with different methods' categories, so they have different behaviors within inconsistency.
	
	\noindent
	\textbf{Dataset.}
	We studied the effect of inconsistency resolution on accuracy using two publicly available datasets for sentiment analysis: News headlines dataset~\cite{cortis2017semeval} and the test data of the sentiment treebank dataset~\cite{socher2013recursive} (sst). 
	To consider the in-tool inconsistency, and for experimental purposes, we augmented the datasets with paraphrases using a generative adversarial network (GAN) \cite{iyyer2018adversarial}.

	For each document in the dataset, we generated three paraphrased documents with the same polarity as the original one.  
    These datasets allow us to study the effect of resolving in-tool and inter-tool inconsistency on accuracy. 
    Note that in our future work, we use a clustering method on the data to create our clusters. 
    
    Statistics about the datasets are presented in  Table~\ref{tab:Statistics_on_datasets}
    
	
\begin{figure}
    \centering
    \begin{minipage}{0.8\linewidth}
        \centering
        \includegraphics[width = 0.99\linewidth]{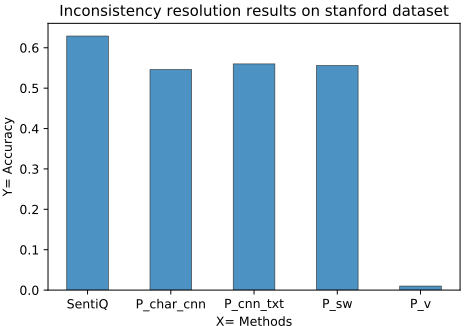}
        
        \caption{Accuracy optimization  on stanfod treebank}
        \label{fig:acc-stanford}
    \end{minipage}
    \begin{minipage}{0.8\linewidth}
        \centering
        \includegraphics[width = 0.99\linewidth ]{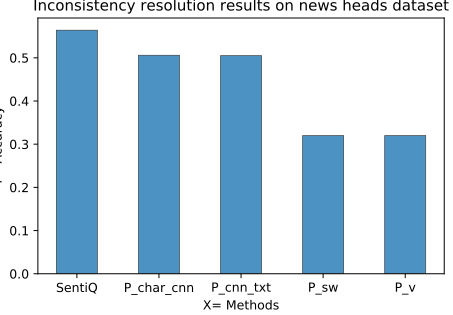}
        \caption{Accuracy optimization  on news headlines}
        \label{fig:acc-news}
    \end{minipage}
\end{figure}

\begin{table}[h]
		\centering
		\scalebox{0.8}
		{%
			\begin{tabu}{|l|c|c|c|c|c|}
				
				\hline
				Statistics & \# elements &  \# Positive&  \# Neutral & \# Negative \\
				\hline
				\makecell{$News\_heads$}  &1583 & 891& 37  & 655\\
								\hline

				\makecell{$SST$}  & 3505 & 1481& 640&1384\\
				\hline
				
		\end{tabu}}
		
		\caption{Statistics on datasets.}
		
		\label{tab:Statistics_on_datasets}
	
	\end{table}

	\noindent
	\textbf{Experiments.}
    To evaluate the efficiency of resolving inconsistencies using SentiQ on the accuracy of the system, we compare it to the Majority Voting (MV) baseline. 
We use MV to resolve the in-tool inconsistency, inter-tool inconsistency, and both inconsistencies; then, we calculate the accuracy on the dataset after resolving contradictions. 
The majority voting for in-tool inconsistency resolution consists of calculating the most repeated polarity in the cluster and attributes it to all cluster documents : 

\noindent
$P_{t_k}(A)= argmax_{\{+,0,-\}} \{ \sum_{d_i \in A}  \mathbbm{1}_{(P_{t_k}(d_i)=+)} , \sum_{d_i \in A}  \mathbbm{1}_{(P_{t_k}(d_i)=0)},$   
$\sum_{d_i \in A}\mathbbm{1}_{(P_{t_k}(d_i)=-)}\}$. Inter-tool inconsistency resolution using majority voting consists of attributing to the document the polarity attributed by most tools:

\noindent
$P_{*}(d_i)= argmax_{\{+,0,-\}} \{ \sum_{P_{t_k} \in \Pi}  \mathbbm{1}_{(P_{t_k}(d_i)=+)} , \sum_{P_{t_k} \in \Pi}  \mathbbm{1}_{(P_{t_k}(d_i)=0)},$   
$\sum_{P_{t_k} \in \Pi}\mathbbm{1}_{(P_{t_k}(d_i)=-)}\}$ 
. Resolving both inconsistencies with MV consists of considering in the cluster all polarities given by polarity functions and attributing to each document the most repeated polarity.

\noindent
\textbf{Accuracy Optimization with SentiQ.}
To evaluate the accuracy improvement obtained by SentiQ, we run SentiQ on the two datasets News headlines and SST.
The Figures~\ref{fig:acc-stanford},\ref{fig:acc-news} present the accuracy of resolving  inconsistencies using SentiQ on the two datasets SST and news headlines respectively with the two queries $IsNegative(d)$ and $IsPositive(d)$ and the polarity functions $P_{char\_cnn}$, $P_{text\_cnn}$, $P_{sw}$ and $P_{v}$.

We observe an accuracy improvement of 0.629 and 0.56  on the two datasets  SST and the news headlines, respectively. These preliminary results prove the efficiency of resolving both in-tool inconsistency and inter-tool inconsistency using SentiQ to improve the accuracy. To analyze the performances and limits of SentiQ, we compare it in the next section to the MV baseline in the presence of variable-sized datasets.

\textbf{Accuracy optimization and dataset size.}
The results are presented in the Table~\ref{tab:acc}. 

We evaluate the accuracy optimization of polarity functions on samples of different sizes ($25$, $100$, $500$ and $1500$) from the news headlines dataset using SentiQ and MV to resolve in-tool inconsistencies, inter-tool inconsistencies, and both of them.  "Original Acc" represents the original accuracy of the polarity function on this dataset, while "MV in-tool" represents the accuracy on different samples after resolving in-tool inconsistency using MV. 
"Inter-tool MV" represents the overall accuracy of the system after solving inter-tool inconsistencies, and the last line of the table represents the accuracy obtained after inferring the polarity of the whole system using our SentiQ.

\begin{table*}[t!]
\centering
		{%
			\begin{tabu}{|l|c|c|c|c|c|c|c|c|} 
				\hline
				\multicolumn{1}{|c|}{Tools} & \multicolumn{4}{c|} {Original Acc} & \multicolumn{4}{c|} {$MV in-tool$} \\ 
				\cline{2-9}
				\multicolumn{1}{|c|}{} &size = 25 &  100 &   500 &  1500& size = 25 & 100 & 500 &  1500 \\
				\hline
				$P_{char\_cnn}$&0.62 &0.55& 0.50&0.506  & 0.69 & 0.59 &0.50&0.52 \\
				\hline
				$P_{cnn\_txt}$& 0.5& 0.48 & 0.52&0.505 & 0.54 &0.54&  {0.57} &0.55  \\
				
				\hline
				$P_{sw}$& 0.34& 0.35 & 0.34&0.33& 0.38 &0.39  & 0.35&0.33    \\
				\hline
				$P_{tb}$ & 0.38 & 0.38 & 0.38&0.40 & 0.46 & 0.41  &0.42 &0.44 \\
				\hline
				$P_{v}$& 0.5  & 0.35 & 0.34&0.0.33&    0.38& 0.39 &0.35&0.33\\
				\hline
				\hline
				$inter\_tool \, MV$ & 0.5  & 0.47 & 0.51& 0.506&   0.42& 0.54 &0.52&{0.515}\\
				\hline
				$SentiQ$ & \textbf{0.76}  & \textbf{0.70} &\textbf{0.60}& \textbf{0.56}&   N/A & N/A  & N/A&N/A \\
				\hline
		\end{tabu}}
		\caption{Accuracy of tools before/after inconsistency resolution. The best performance for each dataset size is marked in bold.}
		\label{tab:acc}
	\end{table*}

\noindent
	\textbf{Results.}
	We observe that resolving in-tool inconsistency increases the accuracy of tools in most of the cases. 
	The only case where we have accuracy degradation corresponds to the tool $P_v$, where accuracy changes from $acc=0.5$ to $acc=0.38$ after resolving inconsistencies. 
	When analyzing the data of this case, we found that most of this tool's predictions where $Neutral$ instead of the data ground truth. 
	As a result, majority voting falsified the results of the correctly predicted instances. 
	
Resolving inter-tool inconsistency using majority voting decreases effectiveness in the case of tools that are incompatible in terms of accuracy (i.e., having widely different accuracy scores). 
Like the case of the two samples of size=25 and size=100 of the Table~\ref{tab:acc}, where the weak nature of $P_v$, $P_{sw}$, and $P_{tb}$ on the datasets has influenced the performance of the voting system (accuracy decreased from $0.62$ to $0.5$ on the dataset of size 25 and from $0.55$ to $0.47$ on the dataset of size 100).  
	SentiQ addresses this problem, because it weighs different tools based on the inconsistencies on the whole dataset. 
    SentiQ provides an accuracy improvement  of $0.76$ on the first dataset, $0.70$ on the second, and $0.60$ on the third dataset, outperforming majority voting.


This leads to other research problems, especially that of scalability, since we could not run experiments with a larger dataset, due to the high inference complexity of the Markov solver. 
Therefore, we need a more efficient Markov logic solver adapted to analyze large scale social media data.
	
	 We also observe that the MLN solver deletes some rules from the model (by attributing a negative, or a $0$ weight), which can penalize the inference.  
	 The final results of the system could be boosted by adding business rules that can improve the polarity inference. 
	 This approach can be applied to various problems such as truth inference in crowd-sourcing, and other classification problems.  We proved that resolving both in-tool and inter-tool inconsistency outperforms using only inter-tool inconsistencies. 

\section{Conclusions and Future Work}
	 In this paper, we presented an MLN-based approach to solve inconsistencies and improve classification accuracy.  
	 Our results show the efficiency of including semantics to resolve in-tool inconsistency. 
	 The initial results of SentiQ are promising and confirm that resolving in-tool inconsistency boosts accuracy. However, to test SentiQ  efficiency in resolving inconsistencies and improving the accuracy of social media data, we need MLN solvers that can scale with the data size. 
	 Finally, we plan to investigate the use of  domain expert rules for improving the polarity inference of SentiQ.
	 
	 \section{Acknowledgement:}
	 This work has been  supported by  the ANRT French program and IMBA consulting.

\bibliographystyle{ACM-Reference-Format}
\bibliography{sample-base}

\appendix

\end{document}